\documentclass[conference]{IEEEtran}
\usepackage{cite}
\usepackage{amsmath,amssymb,amsfonts}
\usepackage{algorithmic}
\usepackage{graphicx}
\usepackage{times}
\usepackage{epsfig}
\usepackage{booktabs}
\usepackage{enumitem}
\usepackage{multirow}
\usepackage{makecell}
\usepackage{color}
\usepackage[dvipsnames, svgnames, x11names]{xcolor} 
\usepackage{stfloats}
\usepackage{textcomp}
\usepackage{algorithmic}
\usepackage[ruled,lined]{algorithm2e}
\usepackage{xcolor}
\def\BibTeX{{\rm B\kern-.05em{\sc i\kern-.025em b}\kern-.08em
    T\kern-.1667em\lower.7ex\hbox{E}\kern-.125emX}}
\usepackage{url}
\usepackage{hyperref}
\usepackage[capitalize]{cleveref}

\begin{document}

\title{XNet v2: Fewer Limitations, Better Results and Greater Universality
% {\scriptsize \textsuperscript{*}Note: Sub-titles are not captured in Xplore and
% should not be used}
% \thanks{Identify applicable funding agency here. If none, delete this.}
}

\author{\IEEEauthorblockN{Yanfeng Zhou\textsuperscript{1,2},
Lingrui Li\textsuperscript{1,2}, Zichen Wang\textsuperscript{1,2}, Guole Liu\textsuperscript{1,2}, Ziwen Liu\textsuperscript{1,2}, Ge Yang\textsuperscript{1,2,*}
}
% \IEEEauthorblockA{\textit{\textsuperscript{1}DAMO Academy, Alibaba Group, Hangzhou, China}}
\IEEEauthorblockA{\textit{\textsuperscript{1}School of Artificial Intelligence, University of Chinese Academy of Sciences, Beijing, China}}
\IEEEauthorblockA{\textit{\textsuperscript{2}Institute of Automation, Chinese Academy of Sciences, Beijing, China}}
\IEEEauthorblockA{\textit{\{zhouyanfeng2020, lilingrui2021, wangzichen2022, guole.liu, ziwen.liu, ge.yang\}@ia.ac.cn}}% <-this % stops an unwanted space
}

% \author{\IEEEauthorblockN{Yanfeng Zhou}
% \IEEEauthorblockN{}
% \IEEEauthorblockA{\textit{School of Artificial Intelligence} \\
% \textit{University of Chinese Academy of Sciences}\\
% Beijing, China \\
% zhouyanfeng2020@ia.ac.cn}
% \and
% \IEEEauthorblockN{Lingrui Li}
% \IEEEauthorblockA{\textit{Institute of Automation} \\
% \textit{Chinese Academy of Sciences}\\
% Beijing, China \\
% lilingrui2021@ia.ac.cn}
% \and
% \IEEEauthorblockN{Ge Yang}
% \IEEEauthorblockA{\textit{Institute of Automation} \\
% \textit{Chinese Academy of Sciences}\\
% Beijing, China \\
% ge.yang@ia.ac.cn}
% \and
% \IEEEauthorblockN{4\textsuperscript{th} Given Name Surname}
% \IEEEauthorblockA{\textit{dept. name of organization (of Aff.)} \\
% \textit{name of organization (of Aff.)}\\
% City, Country \\
% email address or ORCID}
% \and
% \IEEEauthorblockN{5\textsuperscript{th} Given Name Surname}
% \IEEEauthorblockA{\textit{dept. name of organization (of Aff.)} \\
% \textit{name of organization (of Aff.)}\\
% City, Country \\
% email address or ORCID}
% \and
% \IEEEauthorblockN{6\textsuperscript{th} Given Name Surname}
% \IEEEauthorblockA{\textit{dept. name of organization (of Aff.)} \\
% \textit{name of organization (of Aff.)}\\
% City, Country \\
% email address or ORCID}
% }
\maketitle

\begin{abstract}
XNet introduces a wavelet-based X-shaped unified architecture for fully- and semi-supervised biomedical segmentation. So far, however, XNet still faces the limitations, including performance degradation when images lack high-frequency (HF) information, underutilization of raw images and insufficient fusion. To address these issues, we propose XNet v2, a low- and high-frequency complementary model. XNet v2 performs wavelet-based image-level complementary fusion, using fusion results along with raw images inputs three different sub-networks to construct consistency loss. Furthermore, we introduce a feature-level fusion module to enhance the transfer of low-frequency (LF) information and HF information. XNet v2 achieves state-of-the-art in semi-supervised segmentation while maintaining competitve results in fully-supervised learning. More importantly, XNet v2 excels in scenarios where XNet fails. Compared to XNet, XNet v2 exhibits fewer limitations, better results and greater universality. Extensive experiments on three 2D and two 3D datasets demonstrate the effectiveness of XNet v2. Code is available at \url{https://github.com/Yanfeng-Zhou/XNetv2}.
\end{abstract}


\section{Introduction} 
Biomedical image segmentation has achieved remarkable success with the development of deep neural networks (DNNs) \cite{ronneberger2015u,milletari2016v,hatamizadeh2022unetr}. Fully-supervised training is a common learning strategy for semantic segmentation, which is trained with labeled images, using manual annotations as supervision signals to calculate supervised losses with segmentation predictions. Efficient encoder-decoder architecture is the mainstream paradigm in fully-supervised models. This architecture can accurately preserve the boundary information of segmented objects and alleviate overfitting on limited labeled images. Furthermore, some studies extend this architecture to 3D to meet the needs for volumetric segmentation.
% , such as VNet \cite{milletari2016v}, UNet 3D \cite{cciccek20163d}, ConResNet \cite{zhang2020inter}, etc. 
Recently, sequence-to-sequence transformers have become popular for biomedical image segmentation \cite{zheng2021rethinking,cao2022swin,yu2022metaformer}. Some research attempts to combine convolutional neural networks (CNNs) with transformers \cite{xie2021cotr,hatamizadeh2022unetr}, allowing the model to take advantage of both the low computational cost of CNNs and the global receptive field of transformers.

Compared with fully-supervised model with the same number of labeled images, semi-supervised training has superior performance. It learns with a few labeled images and additional unlabeled images, which alleviates the need for laborious and time-consuming annotations. Semi-supervised models use the perturbation consistency of segmentation predictions to construct the unsupervised loss and use it together with the supervised loss of labeled images as supervision signals to guide model training \cite{tarvainen2017mean,ouali2020semi}. Different perturbation strategies and different calculation methods of unsupervised loss produce various semi-supervised segmentation models, such as SASSNet \cite{li2020shape}, MC-Net \cite{wu2021semi}, SPC \cite{Zhou_2023_BMVC}, etc.

\cite{zhou2023xnet} propose an X-shaped network architecture XNet, which can simultaneously achieve fully- and semi-supervised biomedical image segmentation. XNet uses LF and HF images generated by wavelet transform as input, then separately encodes LF and HF features and fuses them. For fully-supervision, XNet extracts and fuses the complete LF and HF information of the raw images, which helps XNet focus on the semantics and details of segmentation objects to achieve higher pixel-wise accuracy and better boundary contours. For semi-supervision, XNet constructs the unsupervised loss based on dual-branch consistency difference. This difference comes from different attention to LF and HF information, which alleviates the learning bias caused by artificial perturbations. 

However, XNet still shows performance degradation when images have little HF information. Furthermore, it also has the limitations in insufficient fusion and underutilization of raw image information.

In this study, we first analyze the limitations of XNet, then make targeted improvements and propose XNet v2. Different from directly using LF and HF images generated by wavelet transform as input, XNet v2 performs image-level complementary fusion of LF and HF images. The fusion results along with the raw images are fed into three different networks (main network, LF network and HF network) to generate segmentation predictions for consistency learning. Furthermore, similar to XNet, we introduce the feature-level fusion modules to better transfer LF and HF information between different networks. XNet v2 achieves state-of-the-art in semi-supervised segmentation while maintaining superior results in fully-supervised learning. More importantly, it still achieves competitive results in some scenarios where XNet cannot work (such as on the ISIC-2017 \cite{codella2018skin} and P-CT \cite{roth2015deeporgan} datasets). Extensive benchmarking on three 2D and two 3D public biomedical datasets demonstrates the effectiveness of XNet v2.

\begin{figure*}[htb]
% \vskip 0.2in
\begin{center}
  % \fbox{\rule{0pt}{2in} \rule{0.9\linewidth}{0pt}}
   \centerline{\includegraphics[width=0.8\linewidth]{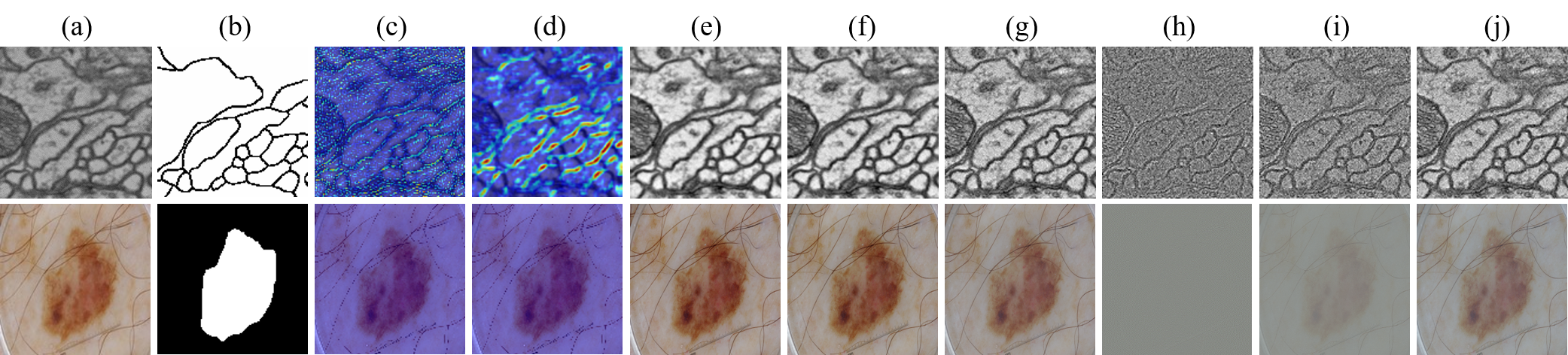}}
       % \vskip -0.1in
   \caption{Comparison of CAM of HF encoder and qualitative show of image-level complementary fusion on CREMI (first row) and ISIC-2017 (second row). (a) Raw image. (b) Ground truth. (c) CAM for the first layer. (d) CAM for the second layer. (e) LF image $I_L$ ($\alpha=0.0$). (f) $x^L$ ($\alpha=0.2$). (g) $x^L$ ($\alpha=0.8$). (h) HF image $I_H$ ($\beta=0.0$). (i) $x^H$ ($\beta=0.2$). (j) $x^H$ ($\beta=0.8$).}
   \label{figure2}
    \end{center}
   \vskip -0.3cm
\end{figure*}

% \textbf{Unified Models with Different Learning Strategies for Semantic Segmentation.} According to the different annotations of training images, learning strategies for semantic segmentation can be divided into four types: fully-, semi-, self- and weakly supervision. The unified models that support multiple learning strategies simultaneously have greater versatility and can be flexibly deployed to practical tasks with different annotation requirements, which makes them more advantageous in situations where annotations are diverse and difficult to acquire. The current unified models have been advanced with the development of DNNs \cite{zhou2023xnet,huang2020self,gao2022segmentation,lee2021anti}. \cite{yao2020saliency} proposes a saliency guided self-attention network for weakly and semi-supervised semantic segmentation. PL-CUT-Seg \cite{moreu2023self} combines synthetic and real images for self- and semi-supervised polyp segmentation. Furthermore, some research attempts to associate semi-supervision with domain adaptation. UMCT \cite{xia2020uncertainty} proposes a unified framework that addresses these two tasks for volumetric medical image segmentation. \cite{hoyer2023improving} uses self-supervised depth estimation to improve semi-supervised and domain adaptive semantic segmentation. \Cref{figure1} shows some versatile models and specialized models in different learning strategies.

\section{Method}
We analyze the limitations of XNet in \Cref{3.1}. Then we propose XNet v2 with fewer limitations and greater universality in \Cref{3.2}. Finally, we further introduce the components of XNet v2, including image-level and feature-level fusion in \Cref{3.3} and \Cref{3.4}, respectively.

\subsection{Limitations of XNet}
\label{3.1}
\textbf{Performance degradation with hardly HF information.} As mentioned in \cite{zhou2023xnet}, XNet is negatively impacted when images hardly have HF information. To intuitively illustrate this phenomenon, we compare the class activation map (CAM) \cite{selvaraju2017grad} of HF encoder of XNet on CREMI \cite{funke2016miccai} and ISIC-2017 \cite{codella2018skin}. From \Cref{figure2}, we can see that CREMI has rich HF information and HF encoder can better focus on these texture and edge details. In contrast, ISIC-2017 has less HF information, which prompts HF encoder to fail to extract recognizable information and locate specific segmentation objects.

\begin{table}[htb]
\vskip -0.15cm
\caption{Comparison of XNet with and without raw images on ISIC-2017 \cite{codella2018skin}. LF+HF+Raw indicates using raw images as additional channels for LF and HF images.}
\label{table_}
\begin{center}
% \begin{small}
\vskip -0.3cm
\begin{scriptsize}
\setlength{\tabcolsep}{1.5mm}
% \resizebox{\linewidth}{!}{
\begin{tabular}{ccc|cccc}
% \begin{tabularx}{\linewidth}{X<{\raggedright}|X<{\centering}X<{\centering}X<{\centering}X<{\centering}|X<{\centering}}
\Xhline{1pt}
Dataset & Method & Input & Jaccard $\uparrow$ & Dice $\uparrow$ & ASD $\downarrow$ & 95HD $\downarrow$\\
\Xhline{1pt}
% \multirow{2}{*}{Fully-} & LF+HF & 79.23 & 88.41 & 0.61 & 3.66 \\
% & LF+HF+Raw & 79.69 & 88.70 & 0.73 & 3.96 \\
\multirow{4}{*}{ISIC-2017} & \multirow{2}{*}{Fully-} & LF+HF & 73.94 & 85.02 & 4.14 & 9.81 \\
& & LF+HF+Raw & \textbf{74.42} & \textbf{85.34} & \textbf{4.11} & \textbf{9.70} \\
\cline{2-7}
% \multirow{2}{*}{Semi-} & LF+HF & 76.28 & 86.54 & 0.76 & 4.19 \\
% & LF+HF+Raw & 77.38 & 87.25 & 1.01 & 5.34 \\
& \multirow{2}{*}{Semi-} & LF+HF & 71.17 & 83.16 & 4.73 & 11.46 \\
& & LF+HF+Raw & \textbf{71.91} & \textbf{83.66} & \textbf{4.53} & \textbf{11.02} \\
\Xhline{1pt}
\end{tabular}
% \end{small}
\end{scriptsize}
\end{center}
\vskip -0.3cm
\end{table}

\textbf{Underutilization of raw image information.} XNet uses LF and HF images generated by wavelet transform as input, and the raw images are not involved in training. Although LF and HF information can be fused into complete information in fusion module, the raw image may still contain useful but unappreciated information. \Cref{table_} compares with and without raw images as input on ISIC-2017 \cite{codella2018skin} and we find that introducing raw images for the dual-branch further improves performance.

\textbf{Insufficient Fusion.} XNet only uses deep features for fusion. Shallow feature fusion and image-level fusion are also necessary. We introduce various fusions for XNet v2. \Cref{table9} and \Cref{table12} in ablation studies of \Cref{4.5} demonstrate their effectiveness.

\begin{figure*}[htb]
% \vskip 0.2in
\begin{center}
   \centerline{\includegraphics[width=0.8\linewidth]{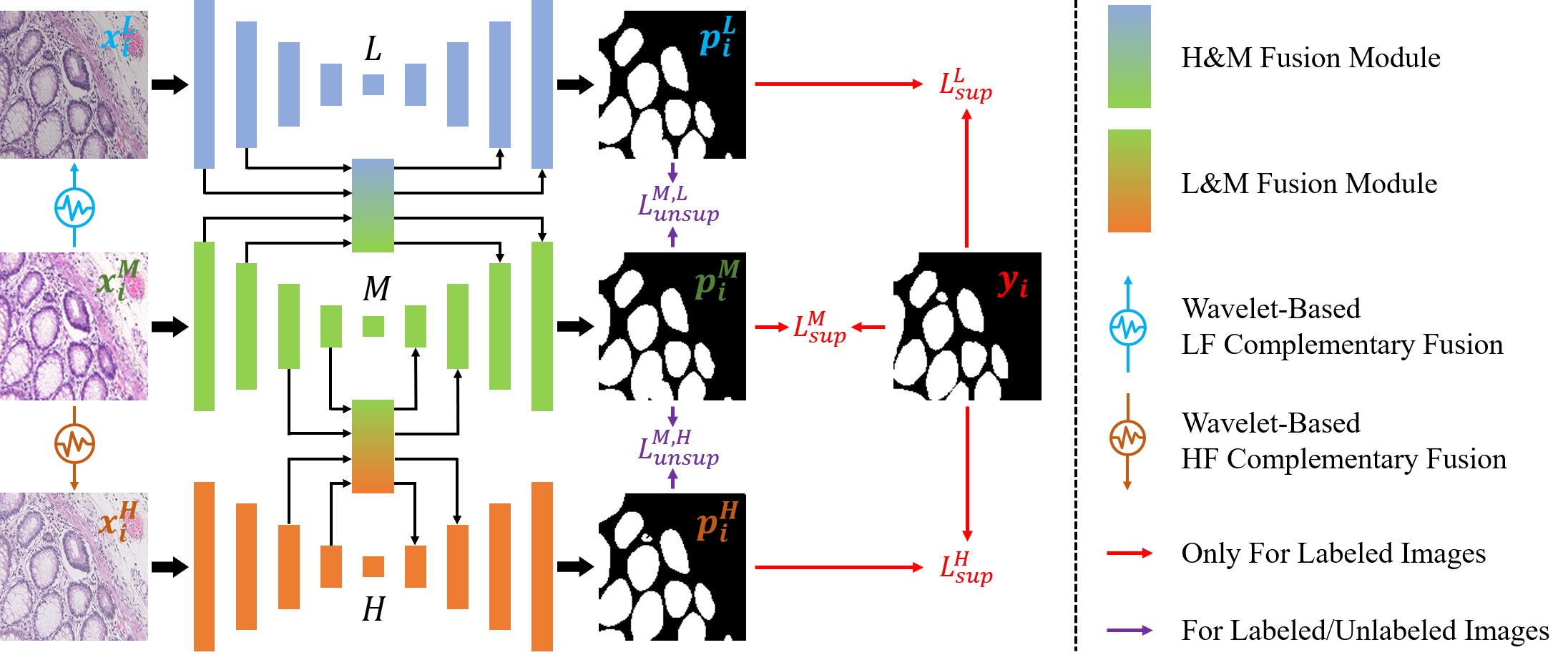}}
   \vskip -0.1in
   \caption{Overview of XNet v2. XNet v2 consists of main network $M$, LF network $L$ and HF network $H$, and uses raw image $x_i^M$, LF complementary fusion image $x_i^L$ and HF complementary fusion image $x_i^H$ as input. XNet v2 learns from unlabeled images by minimizing $L_{unsup}^{M,L}$, $L_{unsup}^{M,H}$, and learns from labeled images by minimizing $L_{sup}^M$, $L_{sup}^L$, $L_{sup}^H$.}
   \label{figure3}
   \end{center}
   \vskip -0.6cm
\end{figure*}

\subsection{Reduce Limitations and Increase Universality}
\label{3.2}
In view of the limitations of XNet, we propose XNet v2 and show its overview in \Cref{figure3}. XNet v2 consists of three sub-networks: main network $M$, LF network $L$ and HF network $H$. $M$, $L$ and $H$ are based on UNet~\cite{ronneberger2015u} (3D UNet~\cite{cciccek20163d}). We use $L$ and $H$ to fuse with $M$ and use their respective shallow and deep features to construct L\&M and H\&M fusion modules, which enables $M$ to better absorb semantics and details. It also allows $L$ and $H$ to generate segmentation predictions with more perturbations. 

Different from directly using LF and HF images generated by wavelet transform as input, XNet v2 performs image-level complementary fusion of LF and HF images, which further reduces limitations and improves universality (we discuss it in detail in \Cref{3.3}). The fusion results along with the raw images are fed into $L$, $H$ and $M$ to generate segmentation predictions for consistency learning.

XNet v2 uses LF and HF outputs to construct consistency loss with the output of $M$ respectively, which avoids the instability of training loss when LF or HF information is insufficient. To be specific, XNet v2 is optimized by minimizing supervised loss on labeled images and triple output complementary consistency loss on unlabeled images. The total loss $L_{total}$ is defined as:
\begin{equation}
  L_{total} = L_{sup} + \lambda L_{unsup},
  \label{eq:1}
\end{equation}
where $L_{sup}$ is supervised loss, $L_{unsup}$ is unsupervised loss, i.e., triple output complementary consistency loss, $\lambda$ is a weight to control the balance between $L_{sup}$ and $L_{unsup}$. Same as \cite{zhou2023xnet}, $\lambda$ increases linearly with training epochs, $\lambda = \lambda_{max}\ast epoch/max\_epoch$. We compare the performance of different $\lambda_{max}$ in ablation studies of \Cref{4.5}.

The supervised loss $L_{sup}$ is defined as:
\begin{equation}
  L_{sup} = L_{sup}^M(p_i^M, y_i) + L_{sup}^L(p_i^L, y_i) + L_{sup}^H(p_i^H, y_i),
  \label{eq:2}
\end{equation}
where $p_i^M$, $p_i^L$ and $p_i^H$ represent segmentation predictions of $M$, $L$ and $H$ for the $i$-$th$ image, respectively. $y_i$ represents ground truth of the $i$-$th$ image. The unsupervised loss $L_{unsup}$ is defined as:
\begin{equation}
  L_{unsup} = L_{unsup}^{M,L}(p_i^M, p_i^L) + L_{unsup}^{M,H}(p_i^M, p_i^H),
  \label{eq:3}
\end{equation}
same as \cite{zhou2023xnet}, $L_{unsup}^{M,L}(\cdot)$ and $L_{unsup}^{M,H}(\cdot)$ are achieved by CPS~\cite{chen2021semi} loss: 
\begin{equation}
\begin{aligned}
  & L_{unsup}^{M,L}(p_i^M, p_i^L)=L(p_i^M, \hat{p}_i^L)+L(p_i^L, \hat{p}_i^M), \\
  & L_{unsup}^{M,H}(p_i^M, p_i^H)=L(p_i^M, \hat{p}_i^H)+L(p_i^H, \hat{p}_i^M),
\end{aligned}
\label{eq:4}
\end{equation}
where $\hat{p}_i^M$, $\hat{p}_i^L$ and $\hat{p}_i^H$ represent pseudo-labels generated by $p_i^M$, $p_i^L$ and $p_i^H$, respectively. As in \cite{zhou2023xnet}, all losses are dice loss \cite{milletari2016v}.

For inference process, we use the segmentation predictions of $M$ as the final result.

% \begin{table*}[b]
% \vskip -0.15in
% \caption{Experimental setup of five datasets.}
% \label{table1}
% \begin{center}
% \vskip -0.1in
% % \begin{small}
% \begin{scriptsize}
% \setlength{\tabcolsep}{3.9mm}
% % \resizebox{\linewidth}{!}{
% \begin{tabular}{cc|ccccc}
% % \begin{tabularx}{\linewidth}{X<{\raggedright}|X<{\centering}X<{\centering}X<{\centering}X<{\centering}|X<{\centering}}
% \Xhline{1pt}
% Dimension & Dataset & Modality & Annotation & \# Train (\# Labeled + \# Unlabeled) & \# Test \\
% \Xhline{1pt}
% \multirow{3}{*}{2D} & GlaS & Optical Microscope & Gland & 85 (17 + 68) & 80 \\
% & CREMI & Electron Microscope & Neuronal Membrane & 3575 (714 + 2861) & 3075 \\
% & ISIC-2017 & Dermoscope & Skin Lesion & 2000 (400 + 1600) & 750 \\
% \hline
% \multirow{2}{*}{3D} & P-CT & CT & Pancreas & 62 (12 + 50) & 20 \\
% & LiTS & CT & Liver and Tumor & 100 (20 + 80) & 31 \\
% \Xhline{1pt}
% \end{tabular}
% % \end{small}
% \end{scriptsize}
% \end{center}
% \vskip -0.15in
% \end{table*}

\subsection{Image-Level Fusion}
\label{3.3}
Different from \cite{zhou2023xnet}, after using wavelet transform to generate LF 
 image $I_L$ and HF image $I_H$, we fuse them in different ratios to generate complementary image $x^L$ and $x^H$. $x^L$ and $x^H$ are defined as:
\begin{equation}
\begin{split}
 x^L = I_L + \alpha I_H, \\
 x^H = I_H + \beta I_L,
\end{split}
\label{eq:5}
\end{equation}
where $\alpha$ and $\beta$ are the weights of $I_H$ and $I_L$, respectively. We can see that the input of XNet is a special case when $\alpha=\beta=0$ while our definition is a more general expression. \Cref{figure2} intuitively compares $x^L$, $x^H$ with different $\alpha$, $\beta$.

\textbf{Simple but Effective.} This strategy is simple but achieves image-level information fusion. More importantly, it solves the limitation of XNet not working with less HF information. To be specific, when hardly have HF information, i.e., $I_H \approx 0$:
\begin{equation}
\begin{aligned}
 & x^L = I_L + \alpha I_H \approx I_L, \\
 & x^H = I_H + \beta I_L \approx \beta I_L \approx \beta x^L.
\end{aligned}
\label{eq:6}
\end{equation}
$x^H$ degenerates into a perturbation form of $x^L$, which can be regarded as consistent learning of raw images with two different LF perturbations. It effectively overcomes the failure to extract features when HF information is scarce.

We set $\alpha$ and $\beta$ to change randomly within the range $[a, b]$ during training stage, which increases the diversity and randomness of training samples to further improve training quality. We compare different range combinations of $\alpha$, $\beta$ and demonstrate the effectiveness of image-level fusion in ablation studies of \Cref{4.5}.

% Furthermore, we have optimized the code - wavelet transform and image-level fusion are now performed synchronously with the data loading process without additional pre-generation operations.

\begin{figure}[htb]
\vskip -0.15in
\begin{center}
\centerline{\includegraphics[width=0.8\linewidth]{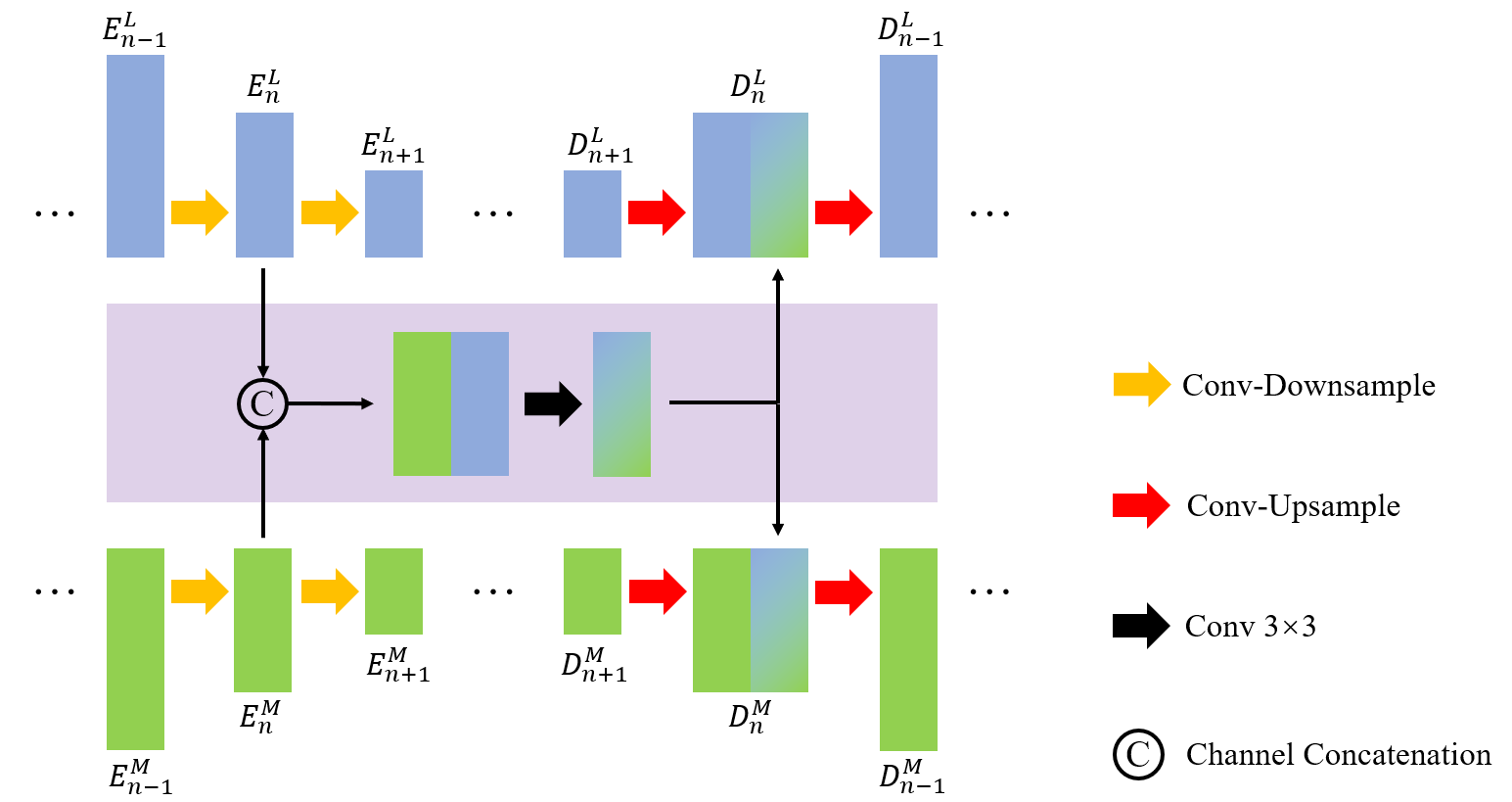}}
      \vskip -0.1in
   \caption{Taking the $n$-$th$ layer features of $M$ and $L$ as an example, visualize the structure of fusion module.}
   \label{figure4}
   \end{center}
   \vskip -0.8cm
\end{figure}

\subsection{Feature-Level Fusion}
\label{3.4}
We use fusion module to transfer feature-level complementary information between $L$ and $M$, $H$ and $M$. Taking L\&M fusion module as an example, we describe its structure. We use $E_n^M$ and $E_n^L$ to represent the $n$-$th$ layer features of $M$ and $L$, respectively. The fusion between $E_n^M$ and $E_n^L$ is shown in~\Cref{figure4}. $E_n^M$ and $E_n^L$ perform channel concatenation to acquire features with twice the number of channels. Then we use 3×3 convolution to fuse features and concatenate the fused features to the decoders of $M$ and $L$. 

For $M$ and $L$, we use deep ($3rd$ and $4th$) features for fusion. For $M$ and $H$, we use shallow ($1st$ and $2nd$) features for fusion. The design of two fusion modules is asymmetric, which is also equivalent to introducing feature-level perturbations into the model.

\section{Experiments}
\subsection{Datasets and Evaluation Metrics}
We evaluate our model on three 2D datasets (\textbf{GlaS} \cite{sirinukunwattana2017gland}, \textbf{CREMI} \cite{funke2016miccai} and \textbf{ISIC-2017} \cite{codella2018skin}) and two 3D datasets (\textbf{P-CT} \cite{roth2015deeporgan} and \textbf{LiTS} \cite{bilic2019liver}). Their preprocessing is the same as~\cite{zhou2023xnet,Zhou_2023_BMVC}.
% and more details for these datasets are shown in \Cref{table1}. 

Following~\cite{zhou2023xnet}, We use Jaccard index (Jaccard), Dice coefficient (Dice), average surface distance (ASD) and 95th percentile Hausdorff distance (95HD) as evaluation metrics.

\subsection{Implementation Details}
We implement our model using PyTorch. Training and inference of all models are preformed on four NVIDIA GeForce RTX3090 GPUs. For 2D datasets (GlaS, CREMI and ISIC-2017), the initial learning rate is set at 0.8. For 3D datasets (P-CT and LiTS), the initial learning rate is set at 0.05. Other experimental setups (such as momentum, training epoch, batch size, training size, etc.) are the same as~\cite{zhou2023xnet}.

\subsection{Comparison with State-of-the-art Models}
\textbf{Semi-Supervision.} We compare XNet v2 extensively with 2D and 3D models on semi-supervised segmentation, including UAMT \cite{yu2019uncertainty}, URPC \cite{luo2022semi}, CT~\cite{CT}, MC-Net+~\cite{wu2022mutual}, etc. From \Cref{table2} and \Cref{table3}, we can see that XNet v2 significantly outperforms previous state-of-the-art models in both 2D and 3D. Furthermore, because of the introduction of image-level complementary fusion and the effective utilization of raw images, XNet v2 has more competitive performance than XNet and is capable of handling scenarios where XNet cannot work (such as on the ISIC-2017 and P-CT datasets), which addresses the limitation of XNet in handling insufficient HF information.

\textbf{Fully-Supervision.} The comparison results are shown in \Cref{table4}. As previous experiments, XNet v2 still shows superior performance compared to XNet.

\begin{table*}[htb]
% \vskip -0.15in
\caption{Comparison with semi-supervised state-of-the-art models on GlaS, CREMI and ISIC-2017 test set. All models are trained with 20\% labeled images and 80\% unlabeled images, which is the common semi-supervised experimental partition. \textbf{\textcolor{red}{Red}} and \textbf{bold} indicate the best and second best performance.}
\label{table2}
\begin{center}
\vskip -0.1in
\begin{scriptsize}
% \begin{small}
\setlength{\tabcolsep}{2.8mm}
% \resizebox{\linewidth}{!}{
% \begin{tabular*}{\hsize}{@{}@{\extracolsep{\fill}}l|l|cccc|cccc@{}}
\begin{tabular}{l|cccc|cccc|cccc} 
\Xhline{1pt}
\multirow{2}{*}{Model} & \multicolumn{4}{c|}{GlaS (17+68)} & \multicolumn{4}{c|}{CREMI (714+2861)} & \multicolumn{4}{c}{ISIC-2017 (400+1600)}\\
& Jaccard $\uparrow$ & Dice $\uparrow$ & ASD $\downarrow$ & 95HD $\downarrow$ & Jaccard $\uparrow$ & Dice $\uparrow$ & ASD $\downarrow$ & 95HD $\downarrow$ & Jaccard $\uparrow$ & Dice $\uparrow$ & ASD $\downarrow$ & 95HD $\downarrow$ \\
\Xhline{1pt}
MT & 76.41 & 86.62 & 2.65 & 13.28 & 75.58 & 86.09 & 1.10 & 5.60 & \textbf{73.04} & \textbf{84.42} & \textbf{4.29} & \textbf{10.53} \\
EM & 76.81 & 86.88 & 2.54 & 12.28 & 73.24 & 84.55 & 1.28 & 6.64 & 70.65 & 82.80 & 4.60 & 11.41 \\
UAMT & 76.55 & 86.72 & 2.73 & 13.43 & 74.04 & 85.08 & 1.10 & 5.71 & 72.55 & 84.09 & 4.37 & 10.70 \\
CCT & 77.60 & 87.39 & 2.27 & 11.23 & 75.74 & 86.20 & 1.31 & 6.93 & 72.80 & 84.26 & 4.35 & 11.12 \\
CPS & 80.46 & 89.17 & 2.08 & 10.56 & 74.87 & 85.63 & 1.25 & 6.47 & 72.42 & 84.00 & 4.39 & 11.55 \\
URPC & 76.84 & 86.91 & 2.31 & 10.97 & 74.70 & 85.52 & \textbf{0.89} & 4.42 & 72.17 & 83.84 & 4.55 & 11.52 \\
CT & 79.02 & 88.28 & 2.33 & 12.02 & 73.43 & 84.68 & 1.23 & 6.33 & 71.75 & 83.55 & 4.56 & 12.17 \\
XNet & \textbf{80.89} & \textbf{89.44} & \textbf{2.07} & \textbf{9.86} & \textbf{76.28} & \textbf{86.54} & \textbf{\textcolor{red}{0.76}} & \textbf{4.19} & 71.17 & 83.16 & 4.73 & 11.46 \\
\hline
XNet v2 & \textbf{\textcolor{red}{83.17}} & \textbf{\textcolor{red}{90.81}} & \textbf{\textcolor{red}{1.75}} & \textbf{\textcolor{red}{8.54}} & \textbf{\textcolor{red}{77.98}} & \textbf{\textcolor{red}{87.63}} & \textbf{\textcolor{red}{0.76}} & \textbf{\textcolor{red}{3.99}} & \textbf{\textcolor{red}{74.07}} & \textbf{\textcolor{red}{85.11}} & \textbf{\textcolor{red}{3.97}} & \textbf{\textcolor{red}{9.95}}\\
\Xhline{1pt}
\end{tabular}
% }
% \end{small}
\end{scriptsize}
\end{center}
% \vskip -0.1in
\end{table*}

\begin{table*}[htb]
\vskip -0.3cm
\caption{Comparison with semi-supervised state-of-the-art models on P-CT and LiTS test set. All models are trained with 20\% labeled images and 80\% unlabeled images. Due to GPU memory limitations, some semi-supervised models using smaller architectures, $\dag$ indicates models are based on lightweight 3D UNet (half of channels). - indicates training failed. \textbf{\textcolor{red}{Red}} and \textbf{bold} indicate the best and second best performance.}
\label{table3}
\begin{center}
\vskip -0.2cm
% \begin{small}
\begin{scriptsize}
\setlength{\tabcolsep}{4.0mm}
% \resizebox{\linewidth}{!}{
% \begin{tabular*}{\hsize}{@{}@{\extracolsep{\fill}}l|l|cccc|cccc@{}}
\begin{tabular}{l|ccccll|cccc} 
\Xcline{1-5}{1pt}
\Xcline{7-11}{1pt}
\multirow{2}{*}{Model} & \multicolumn{4}{c}{P-CT (12+50)} &  & \multirow{2}{*}{Model} & \multicolumn{4}{c}{LiTS (20+80)} \\
& Jaccard $\uparrow$ & Dice $\uparrow$ & ASD $\downarrow$ & 95HD $\downarrow$ & &  & Jaccard $\uparrow$ & Dice $\uparrow$ & ASD $\downarrow$ & 95HD $\downarrow$\\
\Xcline{1-5}{1pt}
\Xcline{7-11}{1pt}
MT & 62.33 & 76.79 & 2.94 & 10.97 & & MT & 72.60 & 80.38 & 10.25 & 27.46 \\
EM & 61.26 & 75.98 & 3.77 & 12.80 & & EM & - & - & - & - \\
UAMT & 62.79 & 77.14 & 3.85 & 14.91 & &  CCT$^{\dag}$ & 73.92 & 81.56 & 11.28 & \textbf{\textcolor{red}{25.03}} \\
SASSNet & 63.67 & 77.81 & 3.06 & 9.15 & & DTC & 74.53 & 82.50 & 12.35 & 35.94 \\
DTC & 64.26 & 78.25 & 2.14 & \textbf{7.17} & & CPS$^{\dag}$ & 71.63 & 79.26 & 9.45 & 28.94 \\
MC-Net & 63.54 & 77.71 & 2.74 & 9.02 & & URPC & - & - & - & -\\
MC-Net+ & \textbf{65.11} & \textbf{78.87} & \textbf{1.89} & 8.15 & & CT$^{\dag}$ & 71.57 & 78.95 & 13.48 & 47.09 \\
XNet 3D & 60.86 & 75.67 & 3.46 & 14.70 & & XNet 3D & \textbf{75.74} & \textbf{83.27} & \textbf{9.26} & 36.88\\
\cline{1-5}
\cline{7-11}
XNet 3D v2 & \textbf{\textcolor{red}{66.96}} & \textbf{\textcolor{red}{80.21}} & \textbf{\textcolor{red}{1.83}} & \textbf{\textcolor{red}{6.31}} & & XNet 3D v2 & \textbf{\textcolor{red}{76.23}} & \textbf{\textcolor{red}{83.92}} & \textbf{\textcolor{red}{8.83}} & \textbf{27.15} \\
\Xcline{1-5}{1pt}
\Xcline{7-11}{1pt}
\end{tabular}
% }
% \end{small}
\end{scriptsize}
\end{center}
\vskip -0.3cm
\end{table*}

\begin{table}[htb]
\vspace{-0.1cm}
\caption{Comparison with fully-supervised models on GlaS, CREMI, ISIC-2017, P-CT and LiTS test set.}
\vspace{-0.15cm}
\label{table4}
% \vskip 0.15in
\begin{center}
% \begin{small}
\begin{scriptsize}
\setlength{\tabcolsep}{1.5mm}
% \resizebox{\linewidth}{!}{
\begin{tabular}{cll|cccc}
% \begin{tabularx}{\linewidth}{X<{\raggedright}|X<{\centering}|X<{\centering}X<{\centering}X<{\centering}X<{\centering}}
\Xhline{1pt}
Dimension & Dataset & Model & Jaccard $\uparrow$ & Dice $\uparrow$ & ASD $\downarrow$ & 95HD $\downarrow$\\
\Xhline{1pt}
% \multirow{3}{*}{GlaS} & UNet & 81.54 & 89.83 & 1.72 & 8.82 \\
\multirow{9}{*}{2D} & \multirow{3}{*}{GlaS} & UNet & 81.54 & 89.83 & 1.72 & 8.82 \\
& & XNet & \textbf{84.77} & \textbf{91.76} & \textbf{1.55} & \textbf{7.87} \\
& & XNet v2 & 84.03 & 91.32 & 1.79 & 9.12 \\
\cline{2-7}
% \multirow{3}{*}{CREMI} & UNet & 75.47 & 86.02 & 1.06 & 5.62 \\
& \multirow{3}{*}{CREMI} & UNet & 75.47 & 86.02 & 1.06 & 5.62 \\
& & XNet & 79.23 & 88.41 & \textbf{0.61} & \textbf{3.66} \\
& & XNet v2 & \textbf{79.80} & \textbf{88.77} & 0.62 & 3.71 \\
\cline{2-7}
% \multirow{3}{*}{ISIC-2017} & UNet & 74.49 & 85.38 & 4.03 & 9.96 \\
& \multirow{3}{*}{ISIC-2017} & UNet & 74.49 & 85.38 & 4.03 & 9.96 \\
& & XNet & 73.94 & 85.02 & 4.14 & 9.81 \\
& & XNet v2 & \textbf{76.04} & \textbf{86.39} & \textbf{3.86} & \textbf{9.78} \\
\hline
% \multirow{3}{*}{P-CT} & UNet 3D & 65.96 & 79.49 & 1.67 & 6.02 \\
\multirow{6}{*}{3D} & \multirow{3}{*}{P-CT} & UNet 3D & 65.96 & 79.49 & 1.67 & 6.02 \\
& & XNet 3D & 70.67 & 82.81 & 1.44 & 5.10 \\
& & XNet v2 3D & \textbf{72.77} & \textbf{84.24} & \textbf{1.40} & \textbf{4.59} \\
\cline{2-7}
% \multirow{3}{*}{LiTS} & UNet 3D & 78.63 & 86.21 & 8.32 & 23.00 \\
& \multirow{3}{*}{LiTS} & UNet 3D & 78.63 & 86.21 & 8.32 & 23.00 \\
& & XNet 3D & \textbf{80.92} & \textbf{87.95} & 5.74 & 18.50 \\
& & XNet v2 3D & 79.53 & 87.09 & \textbf{4.78} & \textbf{16.02} \\
\Xhline{1pt}
\end{tabular}
% \end{small}
\end{scriptsize}
\end{center}
\vskip -0.5cm
\end{table}

\subsection{Ablation Studies}
\label{4.5}
To verify effectiveness of each component, we perform the following ablation studies in semi-supervised learning.

\begin{table}[htb]
\vskip -0.1in
\caption{Comparison of 9 range combinations of $\alpha$ and $\beta$ on GlaS. The wavelet base is Haar.}
\label{table6}
\begin{center}
\vskip -0.1in
% \begin{small}
\begin{scriptsize}
\setlength{\tabcolsep}{1.9mm}
% \resizebox{\linewidth}{!}{
\begin{tabular}{ccc|cccc}
% \begin{tabularx}{\linewidth}{X<{\raggedright}|X<{\centering}X<{\centering}X<{\centering}X<{\centering}|X<{\centering}}
\Xhline{1pt}
Dataset & $\alpha$ & $\beta$ & Jaccard $\uparrow$ & Dice $\uparrow$ & ASD $\downarrow$ & 95HD $\downarrow$\\
\Xhline{1pt}
\multirow{9}{*}{GlaS} & \multirow{3}{*}{$[0.0, 0.4]$} & $[0.0, 0.4]$ & 80.71 & 89.33 & 2.00 & 9.99 \\
& & $[0.2, 0.6]$ & 81.35 & 89.72 & 1.99 & 10.48 \\
& & $[0.4, 0.8]$ & 81.34 & 89.71 & 1.87 & 9.46 \\
% \cline{2-6}
\cline{2-7}
& \multirow{3}{*}{$[0.2, 0.6]$} & $[0.0, 0.4]$ & 80.14 & 88.97 & 2.05 & 10.13 \\
& & $[0.2, 0.6]$ & 81.42 & 89.76 & 1.93 & 9.89 \\
& & $[0.4, 0.8]$ & 81.91 & 90.05 & 1.89 & 9.89 \\
% \cline{2-6}
\cline{2-7}
& \multirow{3}{*}{$[0.4, 0.8]$} & $[0.0, 0.4]$ & 80.57 & 89.24 & 2.02 & 10.40 \\
& & $[0.2, 0.6]$ & 81.10 & 89.56 & 2.02 & 10.26 \\
& & $[0.4, 0.8]$ & \textbf{83.17} & \textbf{90.81} & \textbf{1.75} & \textbf{8.54} \\
\Xhline{1pt}
\end{tabular}
% \end{small}
\end{scriptsize}
\end{center}
\vskip -0.1in
\end{table}

\textbf{Comparison of Range Combinations of $\alpha$ and $\beta$.} As shown in~\Cref{eq:5}, different range combinations of $\alpha$ and $\beta$ produce different LF and HF complementary fusion images. To determine the optimal range combination , we conduct comparative experiments on GlaS. We set 3 value ranges for $\alpha$ and $\beta$ to generate 9 combinations. \Cref{table6} shows the results for GlaS and we find that larger $\alpha$ and $\beta$ achieves better performance. According to the analysis in \Cref{3.3}, this may be because larger $\alpha$ and $\beta$ alleviate the performance degradation of insufficient LF or HF information. 

\begin{table}[htb]
\vskip -0.1in
\caption{Comparison of different trade-off weight $\lambda_{max}$ on five datasets.}
\vskip -0.1in
\label{table7}
\begin{center}
% \begin{small}
\begin{scriptsize}
\setlength{\tabcolsep}{2.0mm}
% \resizebox{\linewidth}{!}{
\begin{tabular}{clc|cccc}
% \begin{tabularx}{\linewidth}{X<{\raggedright}|X<{\centering}|X<{\centering}X<{\centering}X<{\centering}X<{\centering}}
\Xhline{1pt}
Dimension & Dataset & $\lambda_{max}$ & Jaccard $\uparrow$ & Dice $\uparrow$ & ASD $\downarrow$ & 95HD $\downarrow$\\
\Xhline{1pt}
% \multirow{11}{*}{2D} & \multirow{4}{*}{GlaS} & 0.5 & 80.94 & 89.46 & 1.98 & 10.13 \\
\multirow{11}{*}{2D} & \multirow{3}{*}{GlaS} & 1.0 & 81.93 & 90.07 & 1.87 & 9.80 \\
& & 3.0 & 82.14 & 90.19 & 1.82 & 9.39 \\
& & 5.0 & \textbf{83.17} & \textbf{90.81} & \textbf{1.75} & \textbf{8.54} \\
\cline{2-7}
& \multirow{3}{*}{CREMI} & 0.5 & 77.88 & 87.56 & 0.97 & 5.15 \\
& & 1.0 & \textbf{77.98} & \textbf{87.63} & \textbf{0.76} & \textbf{3.99} \\
& & 3.0 & 77.57 & 87.37 & 0.93 & 4.91 \\
% & & 5.0 & 76.51 & 86.69 & 1.09 & 5.77 \\
\cline{2-7}
& \multirow{3}{*}{ISIC-2017} & 1.0 & 74.02 & 85.07 & 4.18 & 11.13 \\
& & 3.0 & 74.07 & 85.11 & \textbf{3.97} & \textbf{9.95} \\
& & 5.0 & \textbf{74.16} & \textbf{85.17} & 4.04 & 10.93 \\
\hline
% \multirow{8}{*}{3D} & \multirow{4}{*}{P-CT} & 0.5 & 65.74 & 79.33 & 2.25 & 7.52 \\
\multirow{6}{*}{3D} & \multirow{3}{*}{P-CT} & 1.0 & 65.90 & 79.45 & 2.15 & 7.19 \\
& & 3.0 & \textbf{66.96} & \textbf{80.21} & \textbf{1.83} & \textbf{6.31} \\
& & 5.0 & 66.85 & 80.13 & 1.89 & 6.89 \\
\cline{2-7}
% & \multirow{4}{*}{LiTS} & 0.08 & 75.16 & 83.08 & 10.23 & 28.07 \\
& \multirow{3}{*}{LiTS} & 0.2 & 75.75 & 83.42 & \textbf{9.16} & \textbf{23.35} \\
& & 0.5 & \textbf{76.27} & \textbf{84.14} & 9.81 & 36.09 \\
& & 1.0 & 74.70 & 82.41 & 9.45 & 32.31 \\
\Xhline{1pt}
\end{tabular}
% \end{tabularx}
% }
% \end{small}
\end{scriptsize}
\end{center}
\vskip -0.1in
\end{table}

\textbf{Comparison of the Trade-off Weight $\lambda_{max}$.} The comparison results of different $\lambda_{max}$ on five datasets are shown in~\Cref{table7}. We find that for relatively easy datasets (GlaS, ISIC-2017 and P-CT), $\lambda$ should increase faster (i.e., $\lambda_{max}$ should large) to highlight the role of many unlabeled images to prevent overfitting. For more difficult datasets (CREMI and LiTS), $\lambda$ should change smoothly (i.e., $\lambda_{max}$ is small), so that the model can better use the labeled images in the early training stage and further improve from unlabeled images in the later training stage.

\begin{table}[htb]
\vskip -0.1in
\caption{Comparison of different perturbations on GlaS. Noise indicates Gaussian noise. INIT indicates network initialization perturbation. SM and SH indicates image smoothimg and sharpening. None indicates without perturbation.}
\vskip -0.1in
\label{table8}
\begin{center}
% \begin{small}
\begin{scriptsize}
\setlength{\tabcolsep}{1.9mm}
% \resizebox{\linewidth}{!}{
\begin{tabular}{ccc|cccc}
% \begin{tabularx}{\linewidth}{X<{\raggedright}|X<{\centering}X<{\centering}X<{\centering}X<{\centering}|X<{\centering}}
\Xhline{1pt}
Dataset & Model & Perturbation & Jaccard $\uparrow$ & Dice $\uparrow$ & ASD $\downarrow$ & 95HD $\downarrow$\\
\Xhline{1pt}
\multirow{10}{*}{GlaS} & \multirow{4}{*}{XNet v2} & Noise & 81.54 & 89.83 & 1.84 & 9.69 \\
& & INIT & 82.08 & 90.16 & 1.88 & 9.91\\
& & SM + SH & 81.43 & 89.77 & 1.92 & 9.80 \\
& & LF + HF & \textbf{83.17} & \textbf{90.81} & \textbf{1.75} & \textbf{8.54} \\
\cline{2-7}
& \multirow{6}{*}{MT} & Noise & 76.41 & 86.62 & 2.65 & 13.28 \\
& & None & 76.57 & 86.73 & 2.71 & 13.72 \\
\cline{3-7}
& & LF & \textbf{77.73} & \textbf{87.47} & \textbf{2.40} & \textbf{11.55} \\
& & SM & 73.37 & 84.64 & 2.74 & 12.88 \\
\cline{3-7}
& & HF & 75.32 & 85.92 & 2.60 & 12.19 \\
& & SH & 67.58 & 80.66 & 3.78 & 18.57 \\
\Xhline{1pt}
\end{tabular}
% \end{small}
\end{scriptsize}
\end{center}
\vskip -0.15in
\end{table}

\textbf{Effectiveness of Wavelet Perturbation.} We compare the wavelet perturbation with other common perturbations in \Cref{table8}, including Gaussian noise, network initialization, image smoothing and sharpening. We find that wavelet perturbation achieved better results. We also note that smoothing and sharpening can also enhance LF semantics and HF details but have a negative impact. Furthermore, we also apply various perturbations to MT \cite{tarvainen2017mean} and acquire consistent conclusions.

\begin{table}[htb]
% \vskip -0.15in
\caption{Comparison of different image-level fusion strategies on GlaS.}
\vskip -0.2in
\label{table9}
\begin{center}
% \begin{small}
\begin{scriptsize}
\setlength{\tabcolsep}{1.9mm}
% \resizebox{\linewidth}{!}{
\begin{tabular}{ccc|cccc}
% \begin{tabularx}{\linewidth}{X<{\raggedright}|X<{\centering}X<{\centering}X<{\centering}X<{\centering}|X<{\centering}}
\Xhline{1pt}
Dataset & $\alpha$ & $\beta$ & Jaccard $\uparrow$ & Dice $\uparrow$ & ASD $\downarrow$ & 95HD $\downarrow$ \\
\Xhline{1pt}
\multirow{4}{*}{GlaS} & 0 (w/o HF) & 0 (w/o LF) & 79.79 & 88.76 & 2.12 & 10.81 \\ 
& 0 & $[0.4, 0.8]$ & 80.18 & 89.00 & 2.22 & 11.87 \\ 
& $[0.4, 0.8]$ & 0 & 79.87 & 88.81 & 2.19 & 11.14 \\
& $[0.4, 0.8]$ & $[0.4, 0.8]$ & \textbf{83.17} & \textbf{90.81} & \textbf{1.75} & \textbf{8.54} \\
\Xhline{1pt}
\end{tabular}
% \end{small}
\end{scriptsize}
\end{center}
\vskip -0.15in
\end{table}

\textbf{Effectiveness of Image-Level Complementary Fusion.} We compare the performance of different image-level fusion strategies, including without fusion ($\alpha\&\beta=0$), single-sided fusion ($\alpha|\beta=0$) and complementary fusion ($\alpha\&\beta\neq0$). From \Cref{table9}, we can see that single-sided fusion hardly has positive effect. It may be because only using single-sided fusion cannot effectively transfer complementary information to the other branch and thus affects the calculation of consistency loss. In contrast, complementary fusion can improve performance by a large margin, because it realizes the mutual complementation of missing frequency information.

\begin{table}[htb]
\vskip -0.1in
\caption{Comparison of different input methods for raw images on GlaS. w/o indicates without raw images as input. Channel indicates inputting raw images into $L$ and $H$ as additional channels of $x^L$ and $x^H$. Branch indicates inputting the raw images into $M$.}
\vskip -0.1in
\label{table10}
\begin{center}
% \begin{small}
\begin{scriptsize}
\setlength{\tabcolsep}{3.3mm}
% \resizebox{\linewidth}{!}{
\begin{tabular}{cc|cccc}
% \begin{tabularx}{\linewidth}{X<{\raggedright}|X<{\centering}X<{\centering}X<{\centering}X<{\centering}|X<{\centering}}
\Xhline{1pt}
Dataset & Raw & Jaccard $\uparrow$ & Dice $\uparrow$ & ASD $\downarrow$ & 95HD $\downarrow$ \\
\Xhline{1pt}
\multirow{3}{*}{GlaS} & w/o & 80.63 & 89.27 & 2.12 & 10.57 \\ 
& Channel & 81.53 & 89.82 & 1.89 & 9.98 \\ 
& Branch & \textbf{83.17} & \textbf{90.81} & \textbf{1.75} & \textbf{8.54} \\
\Xhline{1pt}
\end{tabular}
\end{scriptsize}
% \end{small}
\end{center}
\vskip -0.1in
\end{table}

\textbf{Effectiveness of Raw Images.} As mentioned in \Cref{3.1}, the information of raw images is also crucial for segmentation. In \Cref{table10}, we show the performance improvement of XNet v2 by introducing raw images. Furthermore, introducing additional branches for raw images can further improve performance, so we design additional main network $M$ for raw images in XNet v2.

% \begin{figure*}[htb]
% % \vskip 0.2in
% \begin{center}
%   % \fbox{\rule{0pt}{2in} \rule{0.9\linewidth}{0pt}}
%    \centerline{\includegraphics[width=0.8\linewidth]{figure6.png}}
%       \vskip -0.1in
%    \caption{Qualitative results on GlaS, CREMI and ISIC-2017. (a) Raw images. (b) Ground truth. (c) MT. (d) EM. (e) UAMT. (f) CCT. (g) CPS. (h) URPC. (i) CT. (j) XNet. (k) XNet v2. The blue arrows highlight the difference among of the results.}
%    \label{figure6}
%    \end{center}
%    \vskip -0.3in
% \end{figure*}

% \begin{figure*}[htb]
% % \vskip 0.2in
% \begin{center}
%   % \fbox{\rule{0pt}{2in} \rule{0.9\linewidth}{0pt}}
%    \centerline{\includegraphics[width=0.8\linewidth]{figure9.png}}
%    % \includegraphics[scale=0.7]{figure 1.png}
%    % \setlength{\abovecaptionskip}{-1cm}
%    % \setlength{\belowcaptionskip}{-1cm}
%    % \vspace{-0.05cm}
%          \vskip -0.1in
%    \caption{Qualitative results on P-CT and LiTS. (a) Ground truth. (b) MT. (c) DTC. (d) CPS and MC-Net+. (e) XNet. (f) XNet v2. The blue arrows highlight the difference among of the results.}
%    % \vspace{-0.3cm}
%    \label{figure9}
%       \end{center}
%    \vskip -0.3in
% \end{figure*}

\begin{table}[htb]
\vskip -0.1in
\caption{Comparison of model size and computational cost on GlaS. $^{+}$ indicates to expand the model size by increasing the number of channels. $^{-}$ and $^{--}$ indicate to reduce the number of channels to half and quarter.}
\label{table11}
\begin{center}
\vskip -0.1in
% \begin{small}
\begin{scriptsize}
\setlength{\tabcolsep}{1.4mm}
% \resizebox{\linewidth}{!}{
\begin{tabular}{cl|cc|cccc}
% \begin{tabularx}{\linewidth}{X<{\raggedright}|X<{\centering}X<{\centering}X<{\centering}X<{\centering}|X<{\centering}}
\Xhline{1pt}
Dataset & Model & Params & MACs & Jaccard $\uparrow$ & Dice $\uparrow$ & ASD $\downarrow$ & 95HD $\downarrow$\\
\Xhline{1pt}
\multirow{10}{*}{GlaS} & MT$^{+}$ & 155M & 74G & 76.95 & 86.97 & 2.61 & 13.02 \\
& CCT$^{+}$ & 133M & 77G & 77.80 & 87.52 & 2.32 & 10.61 \\
& URPC$^{+}$ & 116M & 47G & 70.08 & 82.41 & 3.58 & 17.43 \\
& XNet & 326M & 83G & 80.89 & 89.44 & 2.07 & 9.86 \\
& XNet v2 & 113M & 56G & \textbf{83.17} & \textbf{90.81} & \textbf{1.75} & \textbf{8.54} \\
\cline{2-8}
& MT & 69M & 33G & 76.41 & 86.62 & 2.65 & 13.28 \\
& XNet$^{-}$ & 82M & 21G & 79.45 & 88.55 & 2.24 & 11.05 \\
& XNet v2$^{-}$ & 64M & 32G & \textbf{81.49} & \textbf{89.80} & \textbf{1.92} & \textbf{9.52} \\
\cline{2-8}
& XNet$^{--}$ & 20M & 5G & 79.03 & 88.29 & 2.31 & 11.42 \\
& XNet v2$^{--}$ & 7M & 4G & \textbf{81.30} & \textbf{89.68} & \textbf{1.96} & \textbf{9.91} \\
\Xhline{1pt}
\end{tabular}
% \end{small}
\end{scriptsize}
\end{center}
\vskip -0.1in
\end{table}

\textbf{Comparison of Model Size and Computational Cost.} To illustrate that the performance improvement comes from well-designed components rather than the additional parameters brought by multiple networks. We compare the performance of semi-supervised models with the similar scale on GlaS and the results are shown in ~\Cref{table11}. We find that the increase in the number of parameters (Params) and multiply-accumulate operations (MACs) cannot bring positive effects to these semi-supervised models. Furthermore, as in~\cite{zhou2023xnet}, we reduce the number of channels of XNet v2 to half and quarter to generate XNet v2$^{-}$ and XNet v2$^{--}$. These lightweight models still have superior performance than lightweight XNet with similar scale (XNet$^{-}$ and XNet$^{--}$). The above experiments strongly prove that the performance improvement comes from various designs rather than the increase in model size and computational cost.

\begin{table}[htb]
\vskip -0.1in
\caption{Ablation on effectiveness of various components on GlaS, including LF and HF complementary fusion image ($x^{L}$ and $x^{H}$), L\&M and H\&M fusion modules.}
\label{table12}
\begin{center}
\vskip -0.1in
\begin{scriptsize}
% \begin{small}
\setlength{\tabcolsep}{1.2mm}
% \resizebox{\linewidth}{!}{
\begin{tabular}{cccccc|cccc}
% \begin{tabularx}{\linewidth}{X<{\raggedright}|X<{\centering}X<{\centering}X<{\centering}X<{\centering}|X<{\centering}}
\Xhline{1pt}
Dataset & Raw & $x^{L}$ & $x^{H}$ & L\&M & H\&M & Jaccard $\uparrow$ & Dice $\uparrow$ & ASD $\downarrow$ & 95HD $\downarrow$\\
\Xhline{1pt}
\multirow{7}{*}{GlaS} & $\checkmark$ &  &  &  &  & 78.80 & 88.15 & 2.27 & 11.54 \\
& $\checkmark$ & $\checkmark$ &  &  &  & 80.66 & 89.30 & 2.01 & 10.34 \\
& $\checkmark$ &  & $\checkmark$ &  &  & 81.51 & 89.81 & 1.99 & 10.41 \\
& $\checkmark$ & $\checkmark$ & $\checkmark$ &  &  & 81.65 & 89.90 & 1.99 & 10.09 \\
& $\checkmark$ & $\checkmark$ & $\checkmark$ & $\checkmark$ &  & 82.68 & 90.52 & 1.91 & 9.85 \\
& $\checkmark$ & $\checkmark$ & $\checkmark$ &  & $\checkmark$ & 82.11 & 90.18 & 1.90 & 9.63 \\
& $\checkmark$ & $\checkmark$ & $\checkmark$ & $\checkmark$ & $\checkmark$ & \textbf{83.17} & \textbf{90.81} & \textbf{1.75} & \textbf{8.54} \\
\Xhline{1pt}
\end{tabular}
% \end{small}
\end{scriptsize}
\end{center}
\vskip -0.1in
\end{table}

\textbf{Effectiveness of Components.} To demonstrate the improvement of different components, we conduct step-by-step ablation studies on GlaS and the results are shown in \Cref{table12}. Using raw images as input and training the semi-supervised model based on three independent UNet, we achieve a baseline performance of 78.80\% in Jaccard. Using LF and HF complementary fusion images as input improves the baseline by 1.86\% and 2.71\% in Jaccard, respectively. Using them together further improves the baseline to 81.65\% in Jaccard. Introducing L\&M and H\&M fusion modules improves the baseline by 3.88\% and 3.31\% in Jaccard, respectively. By introducing all components, we finally improve the baseline to 83.17\% in Jaccard.

% \subsection{Qualitative Results} 
% \Cref{figure6} and \Cref{figure9} show some qualitative results of different models. By introducing image-level and feature-level HF fusion, XNet v2 achieves less fracture and adhesion of segmentation objects, such as the visual comparison on CREMI and GlaS. By emphasizing LF images and using L\&M fusion module to fuse LF features, XNet v2 can better restore the semantic structure and overall contour of segmentation objects, such as the visual comparison on ISIC-2017, P-CT and LiTS.

\section{Conclusion}
We proposed XNet v2 to solve various problems of XNet, enabling it to maintain superior performance in scenarios where XNet cannot work. XNet v2 has fewer limitations, greater universality, and achieves state-of-the-art performance on three 2D and two 3D biomedical segmentation datasets. Extensive ablation studies demonstrate the effectiveness of various components.

Images are essentially discrete non-stationary signals while wavelet transform can effectively analyze them. We believe that wavelet-based deep neural networks are a novel way for biomedical image segmentation.

\bibliographystyle{IEEEtran}
\bibliography{References}

\end{document}